\documentclass[runningheads]{llncs}
\usepackage[T1]{fontenc}
\usepackage{graphicx}
\usepackage{hyperref}
\usepackage{booktabs}
\usepackage{tabularx} 

\usepackage[backend=biber,style=numeric,sorting=none]{biblatex} 
\begin{document}
\title{Exploring the Potential of Large Language Models for Estimating the Reading Comprehension Question Difficulty\thanks{The manuscript has been accepted for presentation at the 27th International Conference on Human-Computer Interaction in Gothenburg, Sweden, from June 22–27, 2025.}}

%
%
\author{Yoshee Jain\inst{1} \and
John Hollander\inst{2} \and
Amber He\inst{3} \and
Sunny Tang\inst{4} \and
Liang Zhang\inst{5, 6} \thanks{Corresponding Author. Email: lzhang13@memphis.edu} \and
John Sabatini\inst{6, 7}}
\authorrunning{Y. Jain et al.}
%
\institute{Siebel School of Computing and Data Science, University of Illinois Urbana-Champaign, Urbana, IL 61801, USA \\
\and
Department of Psychology and Counseling, Arkansas State University, Jonesboro, AR 72401, USA \and School of Computer Science, Carnegie Mellon University, Pittsburgh, PA 15213, USA \and Heinz College of Information Systems and Public Policy, Carnegie Mellon University, Pittsburgh, PA 15213, USA \and Department of Electrical and Computer Engineering, University of Memphis, Memphis, TN 38152, USA \and Institute for Intelligent Systems, University of Memphis, Memphis, TN 38152, USA \and Department of Psychology, University of Memphis, Memphis, TN 38152, USA}
\maketitle               
\begin{abstract}
Reading comprehension is a key for individual success, yet the assessment of question difficulty remains challenging due to the extensive human annotation and large-scale testing required by traditional methods such as linguistic analysis and Item Response Theory (IRT). While these robust approaches provide valuable insights, their scalability is limited. There is potential for Large Language Models (LLMs) to automate question difficulty estimation; however, this area remains underexplored. Our study investigates the effectiveness of LLMs, specifically OpenAI's GPT-4o and o1, in estimating the difficulty of reading comprehension questions using the Study Aid and Reading Assessment (SARA) dataset. We evaluated both the accuracy of the models in answering comprehension questions and their ability to classify difficulty levels as defined by IRT. The results indicate that, while the models yield difficulty estimates that align meaningfully with derived IRT parameters, there are notable differences in their sensitivity to extreme item characteristics. These findings suggest that LLMs can serve as the scalable method for automated difficulty assessment, particularly in dynamic interactions between learners and Adaptive Instructional Systems (AIS), bridging the gap between traditional psychometric techniques and modern AIS for reading comprehension and paving the way for more adaptive and personalized educational assessments. 
\end{abstract}
%
%
%


\section{Introduction}

Reading comprehension serves as a foundation for interpreting and assimilating written material, empowering people to effectively understand and analyze information \cite{snow2002reading}. Moreover, the implementation of adaptive and effective educational training is pivotal to readers’ success, as it integrates strategies that cultivate robust reading skills and enhance text analysis capabilities \cite{yapp2023effects}. Typically, this skill is evaluated through diagnostic assessments that measure proficiency in pivotal constructs such as sentence construction, vocabulary, and passage-based comprehension, as demonstrated by tools like the Study Aid and Reading Assessment (SARA) system \cite{sabatini2019sara}. Given the increasing complexity of reading tasks, there is a growing need for assessment tools capable of capturing the nuanced interaction between the linguistic and cognitive factors that underlie comprehension. Accurate estimation of question difficulty is essential not only for rigorous assessment but also for informing adaptive learning systems that tailor instruction to individual learner needs \cite{smith2021integrating,psyridou2023reading,mcnamara2002coh}. If questions are too easy, they may fail to challenge learners sufficiently; if too difficult, they can frustrate and hinder learning. Despite this critical need, current methods for automatically assessing question difficulty based on dynamic learner performance remain underdeveloped, limiting their potential to support adaptive instruction and facilitate effective question recommendations in the domain of reading comprehension \cite{alkhuzaey2024text}. 

Traditional approaches to assessing reading comprehension difficulty combine linguistic analysis, using metrics such as Flesch-Kincaid \cite{kincaid1975derivation} and Dale-Chall \cite{chall1995readability}, syntactic complexity and lexical frequency, with psychometric models such as Item Response Theory (IRT) \cite{embretson2013item}. Although these methods reveal key links between textual features and cognitive demand, they rely on extensive human annotation and large-scale pilot testing to capture nuances like semantic depth and inferential demands \cite{snow2008cheap,artstein2017inter}. In contrast, recent machine learning and Natural Language Processing (NLP) techniques automate difficulty estimation by learning feature representations directly from text \cite{collins2014computational}. Desai and Moldovan developed a model using a rich set of features (e.g. rhetorical relations between sentences and syntactic transformations) to classify reading comprehension question difficulty \cite{desai2019towards}. Huang et al. applied word embeddings with an attention-based CNN to predict the difficulty of reading comprehension items \cite{huang2017question}. These approaches reduce the need for large-scale pilot testing and human-driven annotation, thereby offering a scalable and adaptive solution for constructing reading comprehension assessments. This convergence of traditional psychometric techniques with modern AI-driven methods offers a promising pathway toward developing adaptive, data-driven assessment tools that more accurately reflect the multifaceted nature of reading comprehension tasks. This progress can improve the assessment components in adaptive instruction systems for improving reading comprehension. 

Recent advancements in AI, especially Large Language Models (LLMs), have demonstrated remarkable predictive capabilities, including in mathematical reasoning \cite{imani2023mathprompter, ahn2024large} and time series forecasting \cite{jin2023time,gruver2023large,zhang2024large}. Their combined strengths in natural language processing and computing position them as promising tools for assessing question difficulty in reading comprehension. In this study, we investigate the potential of LLMs in identifying question difficulty, to bridge the gap between classical assessment methodologies and modern AI-driven approaches. To this end, our investigation is guided by two primary \textbf{R}esearch \textbf{Q}uestions: 

\begin{itemize}
    \item \textbf{RQ1:} How knowledgeable are LLMs in standardized reading comprehension tasks spanning various subskills (e.g., morphology, syntax, vocabulary, decoding)?
    \item \textbf{RQ2:} To what extent are LLMs capable of recognizing different question difficulty levels?
\end{itemize}

Our study deepens the understanding of LLMs' internal working and highlights their potential to enhance scalable diagnostic tools for both educational assessment and natural language processing. By systematically evaluating LLMs' ability to assess reading comprehension difficulty, we refine automated assessment methods and pave the way for adaptive, personalized learning systems. Ultimately, integrating AI-driven difficulty prediction can optimize instruction, improve student outcomes, and foster more equitable assessment practices across diverse educational contexts. 

\section{Related Work}

This section provides an overview of the primary methods for assessing question difficulty in reading comprehension and examines the application of LLMs in this domain. 

\subsection{Psychometric Assessment with Item Response Theory}

Prior research on question difficulty classification in the domain of reading comprehension relies on traditional psychometric approaches, particularly Item Response Theory (IRT) analysis. IRT provides a mathematical framework for modeling how the latent ability of a test taker and the characteristics of an item (e.g. difficulty, discrimination) interact to produce a response \cite{lord2012applications}. By establishing a statistical relationship between question difficulty and learner ability, IRT facilitates the selection of items that are appropriately challenging for each learner \cite{uto2023difficulty}. Large-scale assessments in reading comprehension, such as the National Assessment of Educational Progress (NAEP), the Programme for International Student Assessment (PISA), and the Progress in International Reading Literacy Study (PIRLS), depend heavily on IRT for calibrating item difficulty and reporting student performance. For instance, the U.S. NAEP employs 2-PL and 3-PL IRT models to estimate both discrimination and difficulty parameters for its reading comprehension items \cite{van2016assessment}. The PISA reading literacy assessment similarly utilizes IRT scaling to summarize student ability and item difficulty across countries \cite{xu2025towards}. The sample-invariance of IRT-derived item difficulties allows test developers to assemble equivalent test forms and maintain a calibrated item bank. Overall, IRT has provided a robust framework for calibrating reading comprehension items, though its dependence on large-scale pilot testing may limit its scalability in dynamic testing environments.

\subsection{Large Language Models for Question Difficulty Estimation}

Recent studies demonstrate that LLMs hold significant promise for predicting difficulty in reading comprehension questions. When employed effectively, these models can detect subtle pitfalls, such as misleading phrasing, or required inference that contributes to the question difficulty. For example, Säuberli et al. used GPT-4 to automatically answer and evaluate multiple-choice items, finding that its assessments of difficulty and quality closely aligned with human evaluators \cite{sauberli2024automatic}. Such research highlights the potential for modern LLMs to serve not only as text generators but also as effective evaluators of question difficulty. Building on these findings, recent work has explored innovative approaches to harness LLMs for difficulty assessment. Park et al. proposed using an LLM as a cohort of ``students'' with varying ability levels to estimate question difficulty \cite{park2024large}. Dutulescu et al. investigated LLM-based measures that were designed to capture the intricate cognitive demands of reading comprehension questions that simpler methods may overlook \cite{dutulescu2024hard}. Notably, prompt design and calibration play crucial roles in these methods; researchers have experimented with strategies such as asking the model to vocalize its reasoning or simulating a student's perspective to yield more reliable difficulty estimates. Xu et al. have introduced adaptive prompting, wherein the style of prompting is adjusted based on the predicted difficulty level (easy versus hard) to elicit optimal reasoning from the LLM \cite{xu2024adaption}. In conclusion, LLM-based approaches offer a scalable and adaptive alternative for estimating question difficulty, complementing traditional methods and paving the way for more dynamic assessment systems. To move towards developing such dynamic and scalable assessments, in this study, we compare the difficulty estimates given by LLMs to the ones retrieved from IRT analysis on questions from the SARA dataset that test reading comprehension proficiency across various skills. With those findings, we gather insights into how these models can complement the traditional psychometric approaches to inform the design of automated difficulty assessment tools.

\section{Method}

In this section, we discuss the dataset used to analyze reading comprehension questions and performance and our method to identify difficulty levels, employing both IRT analysis and LLM models including OpenAI’s GPT-4o and o1. 

\subsection{Dataset}

For this study, we use a dataset with human-generated questions that align with the SARA framework \cite{sabatini2019sara} testing foundational reading skills. This dataset is composed of six subtests that each target a key reading skill. The Word Recognition and Decoding (WRDC) subtest measures both the ability to automatically recognize sight words and to sound out unfamiliar words that are similar to real words, nonwords, and pseudohomophones. The Vocabulary (VOC) subtest assesses understanding of word meanings by requiring students to choose synonyms or meaning associates for academic and content-specific words. In the Morphology (MA) subtest, a cloze format is used to evaluate knowledge of derivational morphology by having students complete sentences with the correct word form. The Sentence Processing (SEN) subtest examines the integration of syntactic and semantic information through cloze items that complete sentences. The Efficiency of Basic Reading Comprehension (EFFIC) subtest employs a maze-format, timed, forced-choice procedure to capture silent reading fluency and basic comprehension at the sentence level. 


The original SARA dataset was collected using a multi-phase, linking design that played a crucial role in its overall methodology. In this design, early phases focused on broad screening and baseline assessments, identifying the key cognitive markers associated with reading ability. Subsequent phases built upon these initial findings with more targeted and detailed assessments, allowing researchers to track developmental changes and establish continuity across different stages of reading development. This linking strategy not only enhanced the efficiency of data collection but also ensured that data from earlier phases could be directly compared with later measurements, thereby providing a rich, longitudinal perspective on the progression of reading skills. Such a design reinforces the robustness of the dataset by enabling cross-validation of constructs over time, ultimately deepening our understanding of how early reading proficiencies predict later literacy outcomes \cite{sabatini2019sara}. Note: Since the study used only secondary data, IRB approval was not required, and it was exempt from human subjects research regulations. 


\begin{table}[h]
    \centering
    \caption{Dataset with Details of the Different Subtests in SARA.}
    \label{tab:performance_metrics}
    \begin{tabularx}{\textwidth}{lXXXXX}  
        \toprule
        & \multicolumn{5}{c}{Subtest} \\
        \cmidrule(lr){2-6}
        Category & WRDC & VOC & MA & SEN & EFFIC \\
        \midrule
        Number of Learners & 2142 & 2142 & 2142 & 2142 & 2142 \\
        Number of Questions & 245 & 206 & 197 & 164 & 196 \\
        \bottomrule
    \end{tabularx}
\end{table}
Table~\ref{tab:performance_metrics} details the number of learners and questions for each reading comprehension subtest in the SARA dataset. A total of 2142 learners participated in all subtests, ensuring that every category had a consistent sample size. The number of questions per subtest varies, reflecting the design of each assessment. 


\subsection{Experiments}

In this section, we present our experimental framework that leverages both traditional psychometric methods and modern LLM-based techniques to assess and classify the difficulty of reading comprehension questions.  

\subsubsection{IRT-based Estimation of Question Difficulty.} 
To establish a benchmark for question difficulty, we use IRT, specifically the 2-Parameter Logistic (2PL) model, to estimate the difficulty and discrimination of each comprehension question \cite{embretson2013item}. The 2PL model includes two key parameters:
\begin{itemize}
    \item \textit{Discrimination parameter $a$}: Captures how well an item differentiates between learners of varying proficiency levels.
    \item \textit{Difficulty parameter $b$}: Measures how challenging an item is, with higher values indicating greater difficulty. 
\end{itemize}

We applied IRT analysis to fit the 2PL model to student response data across all subtests, including WRDC, VOC, MA, SEN, and EFFIC. This procedure estimates both the difficulty and discrimination parameters for each question, thereby providing a direct basis for comparison with predictions generated by LLMs. Consequently, the IRT-derived metrics serve as a robust baseline for evaluating the accuracy of LLMs in classifying question difficulty. 

\subsubsection{LLM Performance Test.}  We first evaluated the performance of two LLMs, GPT-4o and o1, by comparing their ability to answer reading comprehension questions with that of real human students. Accuracy was measured as the percentage of correct responses across all questions. In addition to overall accuracy, we conducted a detailed analysis for each subtest (e.g., WRDC, VOC, MA, SEN, and EFFIC), to identify potential differences in model performance across the distinct skill areas. This granular evaluation enabled us to assess how well each model handled variations in question complexity and content, and to identify specific areas where the models’ performance diverged from human performance. 

\subsubsection{LLM-based Estimation of Question Difficulty.} For the estimation of question difficulty, both LLMs were prompted using a carefully engineered input that simulated the IRT analytical framework. The prompt incorporated the full question text, the accompanying context with multiple-choice options, the ground truth answer, and aggregated student performance data, along with explicit IRT methodological guidelines to steer the estimation process. This ensured that both the baseline and the experimental method were performed using identical contexts, establishing comparability. The temperature setting for both models was fixed at 1 to ensure consistency and reproducibility in response generation. The LLMs then produced estimates of the key IRT parameters, namely, the discrimination parameter ($a$) and the difficulty parameter ($b$), for each question. These LLM-generated estimates were systematically compared to the derived values from the traditional IRT analysis. This approach allowed us to evaluate the viability of LLMs as an alternative or complementary tool for estimating question difficulty. 

\section{Results}

\subsection{Performance Accuracy on Questions from the SARA Dataset}

\begin{figure*}[h!t]
\centering
\includegraphics[width=4.8in]{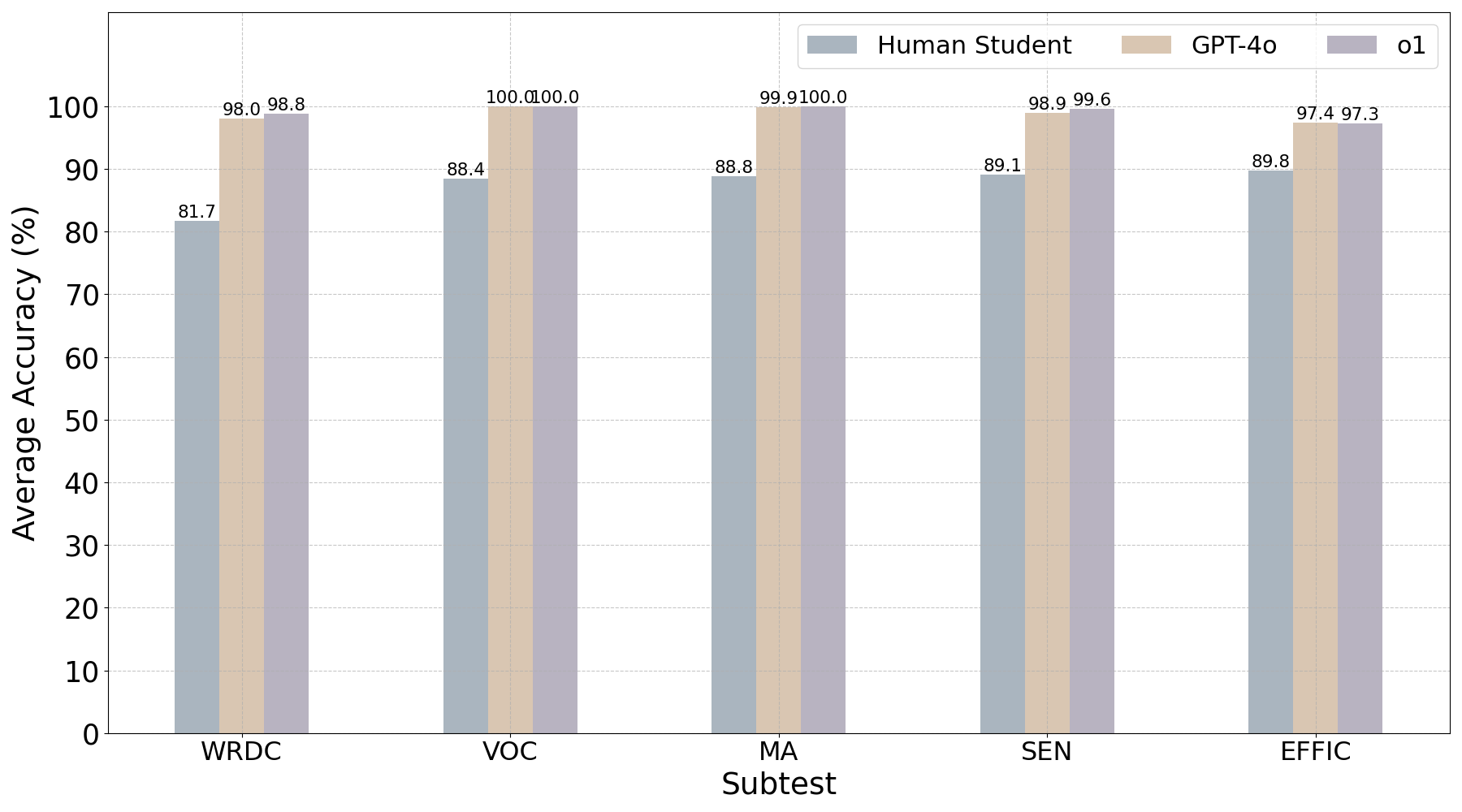}
\caption{Average Accuracy of LLMs for the Different Subtests in the SARA Dataset. Note: WRDC = Word Recognition and Decoding; VOC = Vocabulary; MA = Morphology; SEN = Sentence Processing; EFFIC = Efficiency in Basic Reading. }
\label{fig:comparison_skills}
\end{figure*}

Fig.~\ref{fig:comparison_skills} presents the performance accuracy of LLMs on all subtests from the SARA dataset averaged across three runs. The results indicate that both GPT-4o and o1 models outperform human students. Among human students, the lowest accuracy is recorded in WRDC (81.7\%) and the highest in EFFIC (89.8\%). Notably, both GPT-4o and o1 models achieve near-perfect scores overall, attaining 100\% in VOC (for both models) and in MA (for o1), demonstrating their powerful and robust ability to process nuanced textual information. Overall, o1 outperforms GPT-4o in nearly every subtest, with the only exception being EFFIC, where it trails by a marginal 0.1\%. When averaged across all subtests, human students score 87.57\%, GPT-4o reaches 98.84\%, and o1 achieves 99.14\%, confirming the LLMs' superior performance. These findings imply that even small improvements in reasoning can significantly enhance complex reading comprehension tasks. These results underscore the importance of advanced reasoning capabilities in AI models, particularly evident from o1's performance, in bridging the gap between human and machine comprehension.  Consequently, this evaluation of the SARA dataset provides valuable insights into the problem-solving abilities of large language models and informs future strategies for assessing reading comprehension. 

\subsection{Comparison of the Models in Estimating Question Difficulty}

\begin{figure}[ht!]
\centering
\includegraphics[width=4.8in]{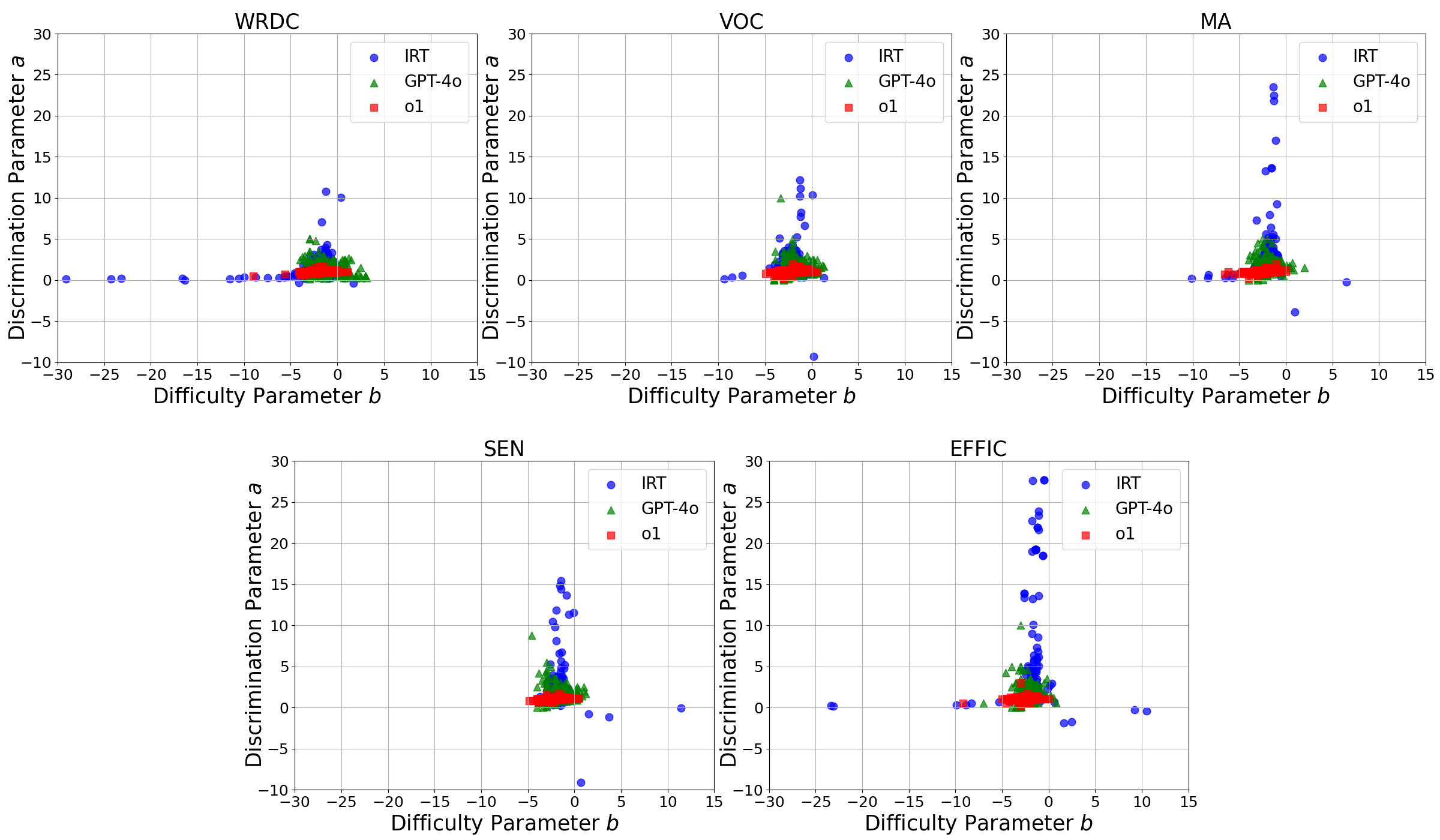}
\caption{Average Accuracy Comparison for Different Subtests in the SARA Dataset.}
\label{fig:comparison_distribution}
\end{figure}

Fig.~\ref{fig:comparison_distribution} shows the distribution for the discrimination parameter \(a\) and difficulty parameter \(b\) for the three experiments with IRT analysis, GPT-4o and o1 models. We note that the IRT model captures a wide range of item complexities grounded in real student responses, while GPT‑4o’s predictions exhibit moderate alignment and cover more variability than the compressed estimates of o1. In particular, the IRT distribution (blue) demonstrates an extensive spread for both parameters, ranging from highly negative difficulty values for easier items to more positive values for challenging ones, and similarly wide-ranging discrimination values that reflect diverse item sensitivities. GPT‑4o’s estimates (green), though generally aligned with the IRT results, tend to cluster within a narrower band, indicating that while they capture the overall trend, they sometimes underrepresent the extreme values observed in the empirical data. Meanwhile, o1’s (red) estimates are notably more compressed, indicating a consistent calibration that emphasizes the central trends in item complexity. This focused distribution provides a stable assessment of difficulty and discrimination, which can be particularly advantageous when a reliable, uniform measure is required for comparative analyses. This comparison highlights the value of integrating LLM-derived estimates with rigorous psychometric analysis to comprehensively capture the subtle characteristics of test question items. 

\subsection{Discussion}

Results from this study demonstrate that LLMs, specifically OpenAI’s o-1 and GPT-4o, not only excel at classifying the difficulty of reading comprehension questions but also outperform human participants across several subtests. These findings carry significant implications for developing automated assessment tools, refining psychometric models, and integrating AI into educational settings. 

This study aimed to address two key research questions. We first investigated the proficiency of LLMs in standardized reading comprehension tasks, encompassing a range of subskills. Fig.~\ref{fig:comparison_skills} highlights a clear advantage of LLMs over human participants in answering questions regarding reading comprehension. Human accuracy ranges from 81.7\% in Word Recognition and Decoding to 89.8\% in Reading Efficiency. In contrast, both LLMs demonstrate near-perfect scores across all subtests, with the o-1 model slightly outperforming GPT-4o in most categories. This suggests that LLMs can process linguistic and cognitive features with exceptional precision, particularly in vocabulary and morphology tasks. The discrepancy between human and machine performance raises important considerations. While LLMs perform well at recognizing and predicting question difficulty, their ability to engage in true comprehension remains debatable. Human participants may struggle with complex word recognition tasks due to cognitive limitations such as working memory constraints or prior knowledge gaps, factors that do not affect LLMs. However, the superior performance of LLMs on factual and rule-based tasks indicates that these models may be less adept at higher-order inference, where nuanced reasoning and contextual understanding are required.

Our second research question examined the extent to which LLMs can recognize different question difficulty levels. From Fig.~\ref{fig:comparison_distribution}, we found that the IRT model captures a wide range of item complexities based on real student responses, with difficulty values spanning from highly negative for easier items to more positive for challenging ones. The GPT-4o model’s predictions show moderate alignment with IRT estimates but tend to cluster within a narrower range. This indicates that while the model can approximate overall trends, it may underestimate extreme values, leading to a less precise representation of question difficulty at both ends of the spectrum. In contrast, the o-1 model’s estimates are more compressed, suggesting a stable yet less variable assessment of difficulty. While this compression may introduce limitations in capturing subtle variations in item complexity, it also provides a more consistent and reliable measure, which could be advantageous for large-scale applications requiring uniform difficulty classifications.

These findings highlight the potential of LLMs to serve as scalable and efficient tools for reading comprehension assessment. By integrating LLM-based difficulty classification with traditional psychometric models, future research can refine automated difficulty prediction to align more closely with human cognitive processes. However, certain challenges remain, particularly in replicating the nuanced reasoning and strategic problem-solving that human readers employ. Additionally, while LLMs demonstrate exceptional accuracy, their reliance on pre-trained statistical associations rather than genuine comprehension suggests the need for further investigation into their interpretative capabilities. Future work could explore hybrid models that integrate LLM-based predictions with cognitive modeling techniques, enhancing their ability to capture the deeper linguistic and conceptual features that contribute to reading comprehension difficulty.

\section{Limitations and Future Work}

Although progress has been made in developing LLM-based assessments for reading comprehension, significant challenges remain. In particular, fully realizing adaptive assessments, where question recommendations adjust to individual learning performance, require further refinement and innovation.

One promising direction is the collection and analysis of detailed reasoning outputs from LLMs. By prompting the models to explain their decisions, we can mine these explanations for insights into the underlying reasoning process. This information is valuable for further prompt optimization and may lead to more accurate assessments. In addition, fine-tuning LLMs based on domain-specific data holds promise for enhancing their performance on targeted tasks.

Another promising avenue is the exploration of multi-agent systems. Collaborative problem-solving among multiple LLMs, as suggested by recent work \cite{latif2024systematic}, could lead to more robust and reliable assessments by combining diverse reasoning strategies. Moreover, integrating these approaches into real-world system scenarios, such as our previously developed SPL system \cite{zhang2024spl}, offers an exciting opportunity to test and optimize LLM-based assessment frameworks in practical settings.

Finally, there are certain computational challenges that need to be addressed. LLMs sometimes struggle with executing complex, iterative machine learning computations due to inherent platform limitations \cite{zhang2024predicting}. Leveraging external computation tools, such as dedicated calculators or specialized ML libraries, may help overcome these computational constraints and improve prediction accuracy.

In summary, while LLM-based assessments for reading comprehension are promising, future work should focus on enhancing prompt design, exploring multi-agent collaboration, integrating real-world testing environments, and augmenting computational capabilities to overcome current limitations. 

\section{Conclusion}

Our investigation demonstrates that LLMs including GPT-4o and o1, can serve as a valuable complement to traditional psychometric techniques for estimating reading comprehension question difficulty. By integrating a 2PL IRT framework with LLM-driven assessments, we found that these models yield difficulty estimates that meaningfully align with empirically derived IRT parameters while also exhibiting differences in sensitivity to extreme item characteristics. Although our findings underscore the potential of LLMs to support scalable, adaptive assessments, they also highlight challenges in achieving computational reproducibility and replicating the nuanced reasoning processes inherent to human comprehension. These insights pave the way for future research aimed at refining prompt engineering, exploring collaborative multi-agent approaches, and incorporating external computational tools to further enhance the precision and applicability of LLM-based evaluations. Ultimately, this work contributes to the development of more efficient and personalized educational assessments that bridge the gap between established psychometric models and modern AI-driven methodologies. 

\section{Acknowledgments}

The research reported here was supported by the Institute of Education Sciences, U.S. Department of Education, through Grant R305F100005 to the Educational Testing Service and R305A190522 to the University of Memphis. The opinions expressed are those of the authors and do not represent views of the Institute or the U.S. Department of Education. 

\printbibliography
\end{document}